\documentclass[12pt]{IEEEtran}
\usepackage[a4paper, left=3.17cm, right=3.17cm, top=2.54cm, bottom=2.54cm]{geometry}
\usepackage[T1]{fontenc}
\usepackage{mathptmx}
\usepackage{amsmath}
\usepackage{amsfonts}
\usepackage{chemformula}
\usepackage{cite}
\usepackage[colorlinks, linkcolor=black, anchorcolor=black, citecolor=black]{hyperref}
\usepackage{graphicx}
\usepackage{listings} 
\usepackage{xcolor}
\usepackage{caption2}
\usepackage{pdfpages} 
\UseRawInputEncoding
\begin{document}
\title{A drone detector with modified
backbone and multiple pyramid feature
maps enhancement structure (MDDPE)}
\author{Chenhao Wu

Maynooth International College of Engineering

Maynooth University
}
\maketitle
\begin{abstract}
    This work presents a drone detector with modified backbone and multiple pyramid feature maps enhancement structure (MDDPE). Novel feature maps improve modules that uses different levels of information to produce more robust and discriminatory features is proposed. These module includes the feature maps supplement function and the feature maps recombination enhancement function.To effectively handle the drone characteristics, auxiliary supervisions that are implemented in the early stages by employing tailored anchors designed are utilized. To further improve the modeling of real drone detection scenarios and initialization of the regressor, an updated anchor matching technique is introduced to match anchors and ground truth drone as closely as feasible. To show the proposed MDDPE's superiority over the most advanced detectors, extensive experiments are carried out using well-known drone detection benchmarks.  
\end{abstract}
    \section{Introduction}
    \par
\begin{figure}[h]
    \centering  
    \includegraphics[width=0.5\linewidth]{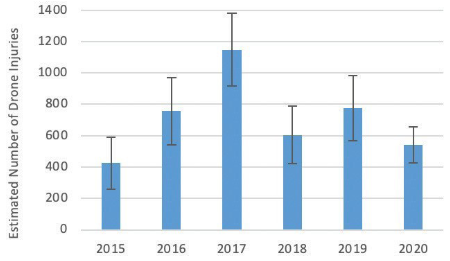} 
    \caption{The statistics about yearly drone-related injuries published by US emergency departments.}
    \label{fig:intro}  
    \end{figure}

 With the development of both hardware devices and software, the current drone or Unmanned Aerial Vehicle (UAV) has undergone both miniaturization and massification. The rapidly prevailing of drones increases a serious number of realistic problems, like threatening public safety, especially airspace safety, or illegal filming.  What's more, the drones also reveal its abilities in military applications. Although most regions have published regulations to prevent the potential threat, the results of drone attacking either by coincidence or on purpose are unaffordable. The Fig.\ref{fig:intro} depicts the data about the yearly number of injuries related to drone in US. It is necessary to build a system to prevent drone attacking and track the friendly drones. A effective system is fundamentally constructed with three part: i- drone detection, classification, and tracking, 
    ii- drone interdiction,
    iii- evidence collection in the case of violation.\cite{overview} As the first stage of the system, a rapidly responding and accurate drone detection is the base of preventing drone attacking. 
    Essentially, drone detection with using visual image can be stored in methodology through image, acoustics or the characteristic signals like raider signals or infrared signatures. Detecting drones from image aspect has unique advantages in both implementation difficulties and cost than other methodologies. Moreover, the scenario of drone detection in real application, usually detecting small objects from both long and close distances in complex background, which makes it difficult to be achieved and generalized by conventional algorithm or manual. However, the machine learning can perform pattern recognition using modalities, which cannot be perceived by humans altogether. These include radio frequencies as well as optic and acoustic signals beyond the abilities of human sense organs.\cite{overview} Visual image-based drone detection has additional advantages as it can provide valuable information about the drone's location, trajectory, and identity. By analyzing the visual features of the drone captured in the image or video, neural network can identify the type of drone, estimate its size and speed, and track its motion over time. This information can be useful for analyzing the drone's behavior and determining potential threats or risks associated with it. Moreover, visual image-based drone detection can be integrated into existing surveillance systems, such as CCTV cameras or aerial monitoring platforms, for enhanced situational awareness and threat assessment. What's more, visual image analysis offers unique benefits in terms of implementation difficulty, cost, and potential for providing valuable information about the detected drone. By leveraging machine learning techniques, visual image-based drone detection can achieve high accuracy and generalizability across diverse scenarios, making it a crucial tool for many applications, including border security, critical infrastructure protection, event monitoring, among others.
    \par The drone detection can be considered as a specific sub-field of object detection with small scale object detection improvement requirements. Fundamentally the common approaches for object detection algorithm in computer vision can be sorted to one-stage algorithm and two-stage algorithms. one-stage algorithms, such as Yolo (you only look once) \cite{yolo} and SSD (single shot detector) \cite{liu2016ssd}, perform detection and classification in a single step. these algorithms use a single neural network to predict the bounding boxes and class probabilities of the objects in the input image. on the other hand, two-stage algorithms, such as faster r-cnn (region-based convolutional neural network) and mask r-cnn, perform detection and classification in two stages. in the first stage, these algorithms generate region proposals that are likely to contain objects of interest. in the second stage, these regions are passed through a classifier to detect and classify the objects accurately.
    \par Although, both of object detection algorithms have been widely improved in both time and accuracy performance, most of improvements are still aimed to enhance the medium and large scale object without loss too much speed. Whereas, the realistic scenario of drone detection also require detecting small objects from far and closed distance with the requirements in both accuracy and time. Although, the existing objects detection including, one-stage and two-stage model, have been improved in a long term, it is still have difficulties to feed the drone detection scenario with the simple implementation of object detection algorithm. 
    \par Since it, it is necessary to develop a accuracy, environment and drone type robustness and high speed drone detection algorithm based on object detection algorithm with using visual image as inputs. Thus, in this work a drone detection algorithm is proposed based on the concept of SSD algorithm. The proposed algorithm use the modified D-Linknet\cite{dlink} with combinatorial architecture of Resnet\cite{restnet} and U-net\cite{uNET} as backbone, multiple enhancement methodology for pyramidal feature maps recombination, tailored prior anchor design for drones' characteristics and specific data augmentation for drone detection scenario stimulation to improve the conventional one-stage methods for a better drone detection performance. In the meanwhile, a mixed dataset is construct based on multiple public dataset and involves to train the model for a better generalization ability and robustness. Ultimately, based on modified D-Linknet and SSD, a drone detection algorithm with multiple pyramid feature maps enhancement structure is formed, which can be called MDDPE. 
    \par Like SSD and Yolo and other object detection algorithms, a standardized and commonly evaluation method, common objects in context (coco) is used in this work to evaluate the performance of algorithm. The coco evaluation standard uses several metrics to evaluate the performance of an object detection algorithm. the most common metric is mean average precision (mAP), which measures the accuracy of the algorithm's predictions by comparing them to ground truth annotations. The precision and recall of the algorithm are calculated based on true positives, and the average precision across all classes is computed. Other metrics used in coco include miss rate, false positive rate, and precision-recall curves. these metrics provide additional information about the algorithm's performance. 
    Besides, a serial of datasets, including Real World\cite{real-world}, Det-fly\cite{Det-Fly}, MIDGARD\cite{MIDGARD}, drone-vs-bird\cite{drone-vs-bird} and USC-drone\cite{usc} are involved to evaluate the performance of to get a comprehensive analysis in different scenario.

    \par In summary, the main contributions of this paper include:
    \begin{itemize}
    \item[$\bullet$] A novel modified backbone to utilize different level information in input image and thus provid more discriminability features for task network.
    \item[$\bullet$] Feature maps enhancement strategies including recollect and reuse the intermediate information in backbone but also utilize different level information in feature extraction network to enhance each other.
    \item[$\bullet$] An tailored improved anchor matching strategy to match anchors and ground truth drone as far as possible to provide better initialization for the regressor.
    \item[$\bullet$]Comprehensive experiments conducted on popular datasets to demonstrate the superiority of our proposed network compared with the state-of-the-art methods and the contributions of each components in our work
    \end{itemize}
   
    \section{Related work}
     As the representative algorithm of two-stage model, RCNN and a series of its improvements\cite{Faster-cnn,he2017mask} achieve a higher accuracy with sacrificing the intolerable time in this scenario. Meanwhile, the one-stage method, representing by Yolo\cite{yolo} and SSD\cite{liu2016ssd} increase the response speed with maintaining a reasonable accuracy by combining prediction and location processes. However, the original one-stage method have a terrible performance in small objects detection. After a serial of works attempting \cite{yolov2,redmon2018yolov3,bochkovskiy2020yolov4,FPN,PAN,li2017fssd}, the performance of one-stage have had a marked climbing in this issue with the contributions from feature map enhancement and various loss function. Since it, it is crucial to determine a suitable one-stage objects detection methodology and improve it to meet the requirements of drone detection.
    The one-stage objects detection model like SSD \cite{liu2016ssd} or Yolo \cite{yolo} achieve a significant higher image processing time with a closed accuracy performance by emerge the region proposals and class classification within a model. In order to fix the shortage, poor ability in distinguishing small group objects and the limitations of input size, of Yolo, Yolo-v2 \cite{yolov2} was proposed with implement the Anchor mechanism and K-means clustering. Continuously, a series of improvements also be proposed in the following versions like Yolov3 \cite{redmon2018yolov3} replace backbone model in Yolov2 or the various tuning tools in Yolov4 \cite{bochkovskiy2020yolov4}. Although the Yolo network have achieved marked performance in detection, the ability in small objects detection of Yolo series is still limited. By comparison, the SSD \cite{liu2016ssd} model include the feature map in all aspects at the beginning which provide a better small object detection performance naturally. The improvement that proposed by SSD itself and other works\cite{PAN,FPN,li2017fssd,fu2017dssd} also provide that the change of initial backbone model, VGG16 to Resent\cite{restnet}, can strength both time and accuracy of SSD model. Beside these, Li et.al\cite{li2017fssd} proposal a fusion feature enhanced backbone to improve the accuracy of SSD further. Inspired by the architecture of U-net, the deconvolution method was implemented in feature map enhancement of DSSD\cite{fu2017dssd}. Similarly, both FPN \cite{FPN} and PAN \cite{PAN} combine different levels of feature maps to improve the detection results. Although the speed and accuracy of both object detection algorithms have significantly improved, the majority of the improvements still focus on enhancing medium and large scale objects without significantly slowing down the speed, which means those algorithm are not suitable to use directly for drone detection.
    \par Most of current implement of drone detection are the specific appliance of existing one-stage model, for instance, Budih et,al\cite{mobilessd} train and verify the ssd model with private datasets and Singha et.al \cite{drone-yolov4} also use private dataset to implement yolo-v4 both of which achieve a good performance in their private dataset. Apart from these, Hu et.al \cite{dorne-improved-yolov3} convert the detection to a quaternary hypothesis test problem than classification image with using cooperative spectrum sensing data at a sensing slot into one image. Kumiawan et.al \cite{yoloplatform} impleted a deep learning platform with using YOLOv2-Tiny, YOLOv3-Tiny, and YOLOv4-Tiny respectively to test the time of yolo network output performance. Corputty et.al replace the kernel size of layers in YOLOv5 to achieve the improvement in drone detection accuracy.\cite{Yolov5_improved_size} The fusion detection methodology was proposals in \cite{anti-UAV} for better drone detection and tracking. What's more, Dadboud et.al \cite{panyolov5} use PANet and mosiac augmentation in YOLOv5 achieving the marked improvement in both drone detection and classification. \cite{droneMaskcnn}, based on the two stage object detection algorithm, Mask RCNN, implemented a high performance drone detection model in map.
    
    \par The dataset of drone detection is also crucial to allocate the features of drone and verify the performance of model within a standard case. As pervious mentioned, the shot view, scale, distance and background of detecting drone are various. Pawełczyk et,al\cite{real-world} collect drone images from video in Youtube and taking to form Real World. Although the views and the resolution of image is limited, the real world dataset have the most types of drone and backgrounds. Det-Fly\cite{Det-Fly} collect drone image from other aircraft which allow the comprehensive views of drone. Meanwhile, the MIDGARD\cite{MIDGARD} and Drone-vs-Bird\cite{drone-vs-bird} include other air objects like birds which allow the research on the problem related to the dingutishing of birds and drone in long distance detection. 
    \section{The problem}
    The problem that encounter in this work can be essentially sorted to two parts: the difficulties during model improvement and the gap between different datasets.
    As previous mentioned, although both one-stage and two-stage algorithm have been significant improvements in both the speed and accuracy of object detection algorithms, these improvements have mainly focused on enhancing medium and large-scale objects without significantly slowing down the speed. However, these algorithms may not be suitable for direct use in drone detection due to their inability to effectively detect smaller drones without reducing the algorithm's speed. The problem with using current object detection algorithms lies in finding a balance between accuracy and time performance for drone detection. While some algorithms may achieve high accuracy, they often require considerable computational resources and time to process large amounts of data, making them impractical for real-time applications such as drone detection. On the other hand, faster algorithms like Yolo algorithm, that prioritize speed over accuracy may miss detecting small drones, leading to false negatives.
    \par The datasets for drone detection have significant differences in terms of the types of drones, backgrounds, and viewing angles. The Real World dataset contains the most types of drones and backgrounds but has a limited resolution and mostly top-down and up-close views, which restricts its use for detecting drones from a bird's-eye view. The Det-Fly dataset overcomes this limitation by capturing multiple drone poses from different viewing angles, but it only includes one type of drone, making it inadequate for detecting other types. The MIDGARD and USC-Drone datasets suffer from a lack of diversity in drone types and viewing angles. If a drone detection algorithm is developed using only one dataset, it may not be able to detect all types of drones and in all scenarios leading to limited performance and accuracy. 
    \par In addition to the differences in types of drones, backgrounds, and viewing angles, another gap between datasets for drone detection lies in their size. The drone detection algorithm typically resizes these images to a fixed size such as 512*512, which may cause a loss of important features for large images. This can result in reduced accuracy and performance of the detection algorithm when applied to larger drone images.
    
    \section{The solution}
    \par In this work, four unique solutions are presented to each of the aforementioned problems. First, we depict a modified states-of-art segmentation model, D-Linknet\cite{dlink}, as our work backbone to enhance the discriminability and information extraction abilities of our work, which combine the concept of Resnet\cite{restnet} and U-net\cite{uNET}. Secondly, inspired by FPN \cite{FPN}, PAN\cite{PAN} and yolo-v5\cite{panyolov5}, we proposed our feature maps enhancement methodology, including features maps supplement function and feature maps recombination enhancement, for a information enhancement extracted feature maps and a better prediction results. Specifically, we assign tailored anchor sizes and scale for default boxes with the analysis on the characteristics of drone and the inheritance of SSD\cite{liu2016ssd} concept. Third, we propose a Tailored Improved Anchor Matching (TIAM), which incorporates anchor partition technique and anchor-based data augmentation to improve the alignment between the ground truth drones and the anchors, resulting in superior initialization for the regressor. The four components are compatible with one another, allowing these strategies to cooperate to enhance performance even further.
     \par In this section, the pipeline of proposed drone detection framework will be firstly introduced, and then detailly elaborate the backbone selection and its modification for model in Sec.\ref{Backbone selection and Modification}, followed by feature maps enhancement functions in Sec.\ref{Feature supplement model} and Sec.\ref{Feature maps recombination enhancement}, sequentially. After that, Sec.\ref{Tailored prior anchor design matching and loss} depicts the scale, anchor and tailored prior anchor matching design of our work.  Sec.\ref{Implementation Details} elaborates the parameter setting during the training. 
\begin{figure}[t]
  \centering  
  \includegraphics[width=\linewidth]{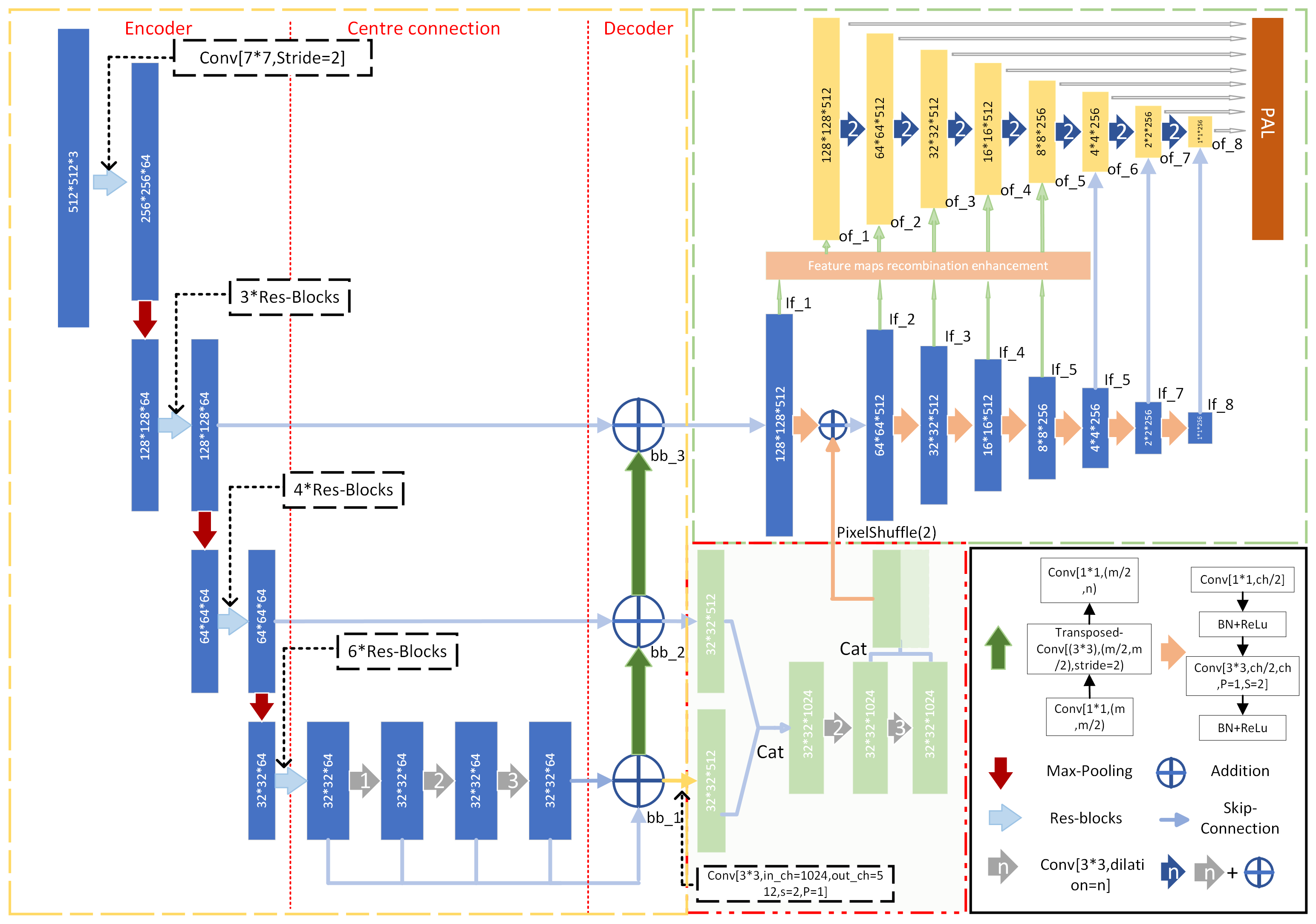} 
  \caption{The framework of MDDPE uses a modified D-Linknet Backbone in yellow frame, Feature supplement Function to provide supplement feature information in red frame, the green frame depicts the feature maps function and its enhancement methodology and legend in black solid cable frame.}  
  \label{fig:framework}  
\end{figure}  
    \subsection{Pipeline of model}\label{Pipeline of model}
    The framework of this work is depicted as Fig.\ref{fig:framework} shown. In this work we involve the concept of D-linkNet\cite{dlink}, a original road extraction model, in our architecture with a serious of truncation, modifications and adding some ameliorated structure to form a specific backbone for this work. Following the concept of one-stage detection algorithm like SSD\cite{liu2016ssd} and Yolo\cite{yolo}, a series of feature maps, whose sizes decline by half from $128*128$ to $1*1$, will be generation for drone features extraction with a complementary enhancement information from segmentation network at beginning stage. By assimilating the feature maps enhancement methodology from FPN\cite{FPN}, PAN\cite{PAN}, FSSD\cite{li2017fssd}, DSSD\cite{fu2017dssd} and a series of Yolo\cite{yolo,yolov2,redmon2018yolov3,bochkovskiy2020yolov4}, the feature maps recombination enhancement are proposed and implemented in the MDDPE. With the analysis about features of drone in different detection scenario, a hierarchical weight scale design is proposed to approach drone features in difference cases. Beside, the tailored anchor matching strategy, including target transformation and specific feature ablation, and mish activation function provide a comprehensive and generalized understanding of model. For loss function, the MDDPE inherit the loss design from SSD, using a weighted sum of the localization loss (loc) and the confidence loss (conf). Ultimately, a more flexible and robustness drone detection model is constructed with higher drone recall and precision in different scales.

\subsection{Backbone Selection and Modification}\label{Backbone selection and Modification}
\begin{figure}[h]
  \centering  
  \includegraphics[width=\linewidth, height =0.4 \linewidth]{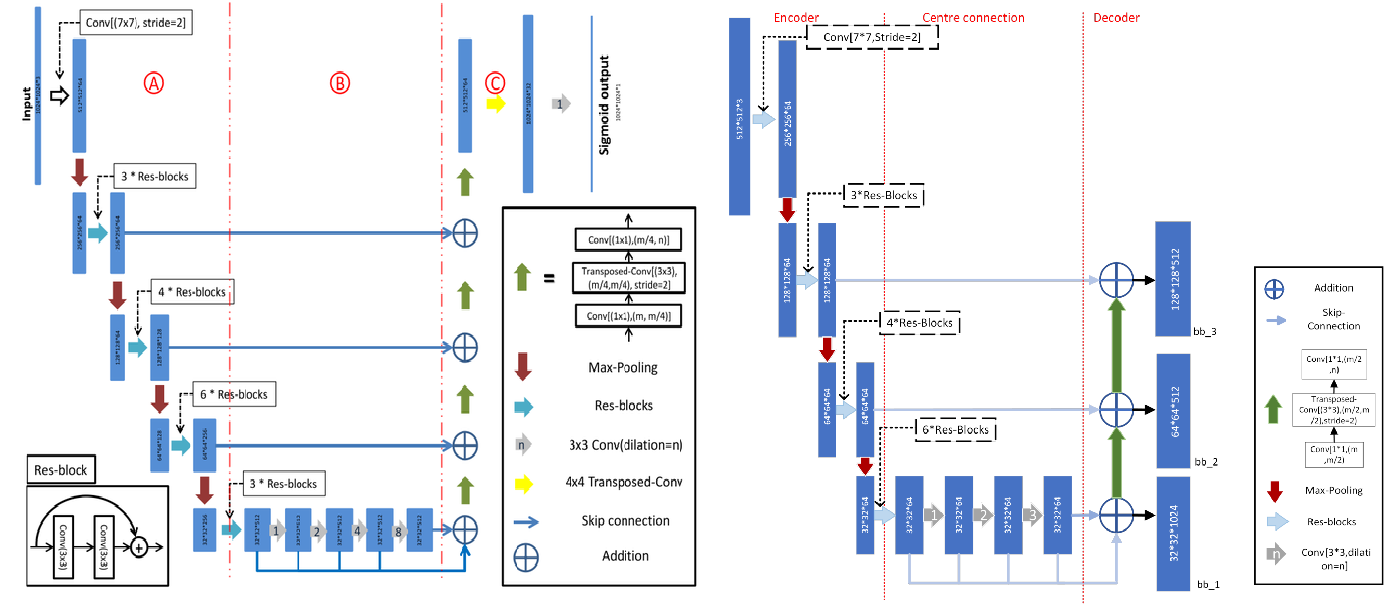} 
  \caption{The architecture of D-linkNet(left) and the modified backbone based on D-LinkNet(right)\\ i.e The A, B, C in left represent encoder,  centre connection and decoder respectively.}  
  \label{fig:backbonecompare}  
\end{figure} 
    Conventionally, the object detection algorithm can be functionally considered as the combination of segmentation network and task network. Although the performance VGG16, Resnet\cite{restnet} or Darknet has been widely proved in various kinds of object detection algorithms, the comprehensive drone features extraction, especially for small scales information, still will be limited by the original squeezed feature map come from these backbones. As a state-of-art segmentation network, D-linkNet\cite{dlink}  adopted encoder-decoder structure by combining the concept of U-net\cite{uNET} and Resnet\cite{restnet}. Besides, the dilated convolution in the center also give the backbone abilities to enlarge the receptive field offeature points without reducing the resolution of the feature maps. By using the transposed Convolution in decoder part, the task network is permitted to extract more detail features from a larger original feature maps. Like other object detection models, in this work, the original D-linkNet is truncated, modified and added some ameliorated structure to form a backbone for our drone detection model. 
  \begin{figure}[h]
  \centering  
  \includegraphics[width=\linewidth]{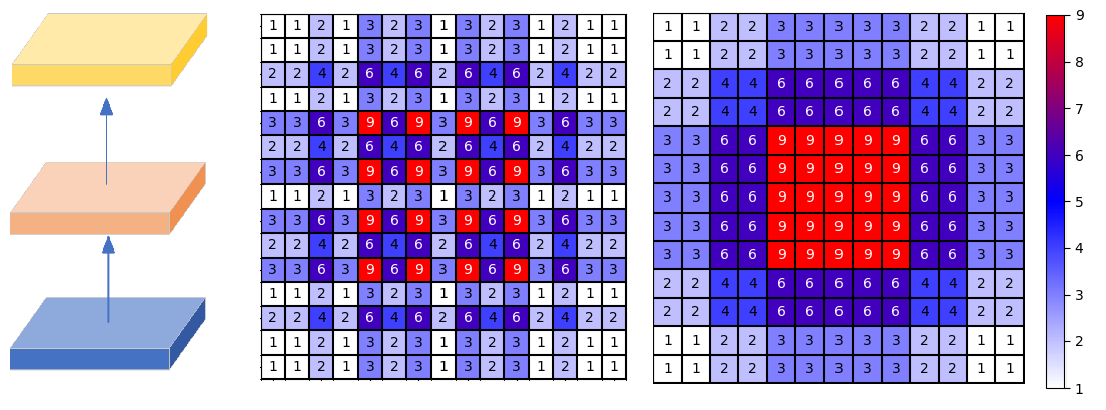} 
  \caption{The map of recall on original matrix after a serial dilated convolution with dilation setting in [1,2,4](left) and [1,2,3](right)}  
  \label{fig:dilations}  
\end{figure} 
    \par The Fig.\ref{fig:backbonecompare} compare the architecture proposed by D-linkNet and the proposed backbone, modifying based on D-linkNet. Unlike other works did, for the purpose of reducing the influences from transposed Convolution and too much task network input parameters, we not only D-linkNet cut prior to classification layers but also wipe off the last decoder layer and the following layers. Thus, even with giving a narrower input size, $512*512$, the backbone model is able to generate the first larger and informative feature map, whose size is $128*128$. Moreover, as Fig.\ref{fig:backbonecompare} shown, we discard the bottom encode-decoder connection with transport the center dilated convolution layers to the upper encode-decoder connection and choose Resnet-50 as the encoder instead of Resnet-34. 

    \par The Fig.\ref{fig:dilations} depict the analysis of the circumstances of matrix recall by the dilated convolution setting in D-linkNet and modified dilated convolution setting in this work. It is notably that the recall on original matrix by original serial dilated convolutions is separated, whereas the new setting form a comprehensive and jagged extractor avoiding information lost in serial dilated convolutions. Our new dilated convolution parameters are accorded with Hybrid Dilated Convolution(HDC) principles and satisfied the function:
 \begin{equation}
 M_i = max [ M_{i+1} - 2r_i, M_{i+1} - 2(M_{i+1} - r_i), r_i ]
\label{eq:dilation-requires}
\end{equation}
proposed in\cite{wang2017understanding}, with $M_n = r_n$ and $M_2 \leq K$. Ultimately, a backbone with reasonable depth and accuracy is formed.
\subsection{Feature maps supplement Function}\label{Feature supplement model}
    The Feature supplement function is aimed to provide more original information within backbone. Although the transposed convolution generate a extraordinary upsampling result, the sequential transposed convolutions usually cause chessboard effect or the distortion of information. Feature supplement Function utilizes the different encoder-decoder layers' results to generate a supplement feature map to enhance the feature map extraction. Our strategy use the concept of dense upsampling convolution (DNC) to upsample the layer in backbone for a enhancement layer. Specially, the enhanced feature map,  $f_{supplement~feature~map}$, can be mathematically defined as follow:
\begin{equation}
\begin{split}
  f_{supplement~feature~map} = f_{PixelShuffle(2)}(f_{concat}(Submap_{(i,j,1)},Submap\_2_{(i,j,2)})
  \label{eq:total}
\end{split}
\end{equation}
where:
\begin{equation}
\begin{split}
  Submap_{(i,j,l)} =f_{serial~dilated~conv}^{n}(f_{concat}(bb\_ 1_{(i,j)},f_{dilated ~conv})(bb\_2_{(i,j)}))
  \label{eq:total}
\end{split}
\end{equation}
    The $(i,j,l)$ signifies a cell that is positioned in the $l$-th layer's feature maps' $(i,j)$ coordinates, and $f$ stands for a set of fundamental operations such upsampling, concatenation, elem-wise generation, and dilation convolution, whose up symbol $n$ stand for serial dilation convolution. In the meantime, the $Submap$ is the feature map after $n+1$ times dilation convolution and the $bb\_n$ is the layer in backbone as shown in Fig\ref{fig:framework}. Finally the supplement feature map will be add after feature extraction operations on first originally feature map.
\subsection{Feature maps recombination enhancement}\label{Feature maps recombination enhancement}
    Feature maps recombination enhancement is able to enhance original features to make them more discriminable and robust, which is called FMRE for short. Inspired by FPN\cite{FPN}, PAN\cite{PAN} and a series improvement of Yolo\cite{yolo,yolov2,redmon2018yolov3,bochkovskiy2020yolov4,panyolov5}, our FMRE also use the concept of feature map enhancement from bottom to up followed by up to bottom. It is worthy that, the most of enhancement is add on original feature maps, which are extracted in SSD\cite{liu2016ssd} proposed methodology, so that our model have abilities to save memory and computing resource during the prediction. After two direction enhancement paths, all feature map will be sent into a $3*3$ conventional layer and normalization layer for a better prediction.
    \par
\begin{figure}[h]
  \centering  
  \includegraphics[width=0.5\linewidth]{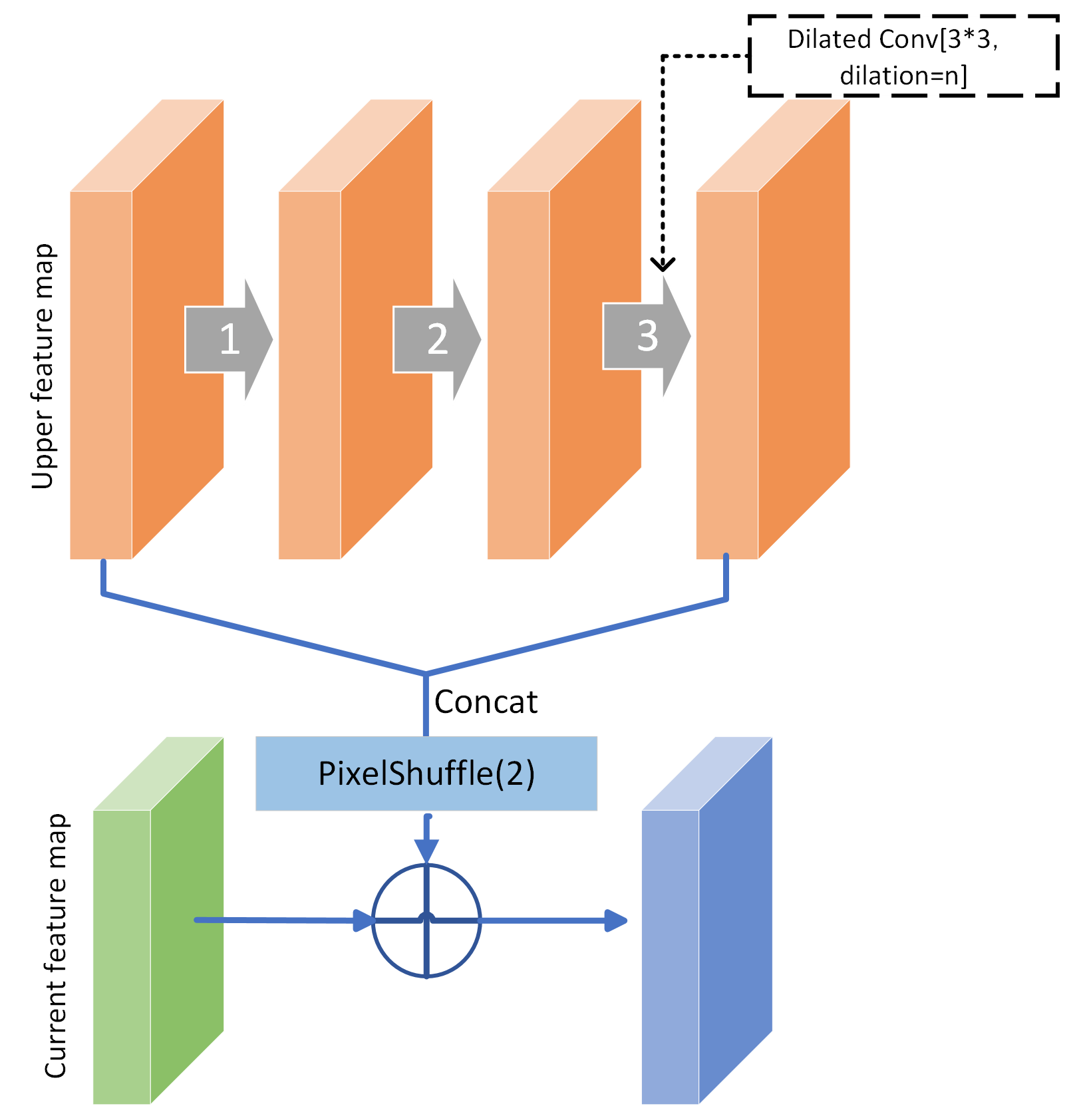} 
  \caption{The architecture of a up to bottom model in Feature maps recombination enhancement}
  \label{fig:mapenhance}  
\end{figure}

 For up to bottom enhancement, we abnegate the methodology that proposed in PAN\cite{PAN} and FPN\cite{FPN}. Although conventional upsampling for upper layer and adding it to lower layer is simple and can save the calculation resources. However, the limitations and potential problems of the conventional upsampling make it may not suitable for the feature maps of drone detection, for example, the nearest neighbor upsampling usually create a highly pixelated, low quality image and bilinear upsampling can cause blurring of sharp edges or lines in an image due to the averaging process. Moreover, the framework of PAN\cite{PAN} and FPN\cite{FPN} involve the depth declining of feature maps which make maps have less capacity to capture complex patterns, contain less high-level abstractions and not be able to generalize well for new data. Since it, we adopt the concept of super-resolution, which usually use a neural network to predict high-resolution images from low-resolution images. As Fig. \ref{fig:mapenhance} shown we use a serious of dilated convolution design to generate there maps, whose size is same to the upper feature maps, which is accorded with HDC principles. After that, the four maps will be cat together and use a periodic sampling in DNC in order to create a twice size maps to add with the lower maps.
    \par For bottom to up enhancement, we absorb the down sampling methodology to enhance the upper feature maps. Unlike the method that use in PAN\cite{PAN} and Yolo-v5\cite{panyolov5}, we replace the max-pool method to avoid potential problem of these methods like potential loss of location information and reduced sensitivity to small-scale features. By using a dilated convolution with a larger receptive field to extract feature on lower feature maps, achieving the purposes of down sampling. Thus the enhanced upper feature maps with using lower layer information, $btuec_{i,j,l}$, can be represented as:
\begin{equation}
\begin{split}
  btuec_{(i,j,l)} =fm_{(i,j,l)}+f_{dilated~conv2}(fm_{(i,j,l-1)})
  \label{eq:total}
\end{split}
\end{equation}
    where, $(i,j,l)$ denotes a cell that is located in the $(i,j)$ coordinates of the feature maps ($fm$) of the $l$-th layer, and f denotes a collection of basic operations. Prior sent to prediction convention layer, all feature maps are going to be normalizated with a mish activation function and a $3*3$ convention layer to align the potential numerical explosion.

\subsection{Tailored prior anchor design and anchors matching portfolio}\label{Tailored prior anchor design matching and loss}
    In order to identify which anchor corresponds to a drone's bounding box during training, positive and negative anchors must be computed. The ground-truth drone and the anchor are currently matched using a bidirectional technique. As a result, for improved initialization of the regressor, anchor design and drone sampling are combined during augmentation to match the anchors and drones as closely as possible. Each feature map cell's association with the fixed form anchor is shown in detail in Table \ref{table:sacle} along with other anchor design elements. Originally, the scale of prior predict default boxes, $s_k$, follows the equation in SSD\cite{liu2016ssd} as follow. By assimilating the concept of SSD\cite{liu2016ssd}, FSSD\cite{li2017fssd}, DSSD\cite{fu2017dssd}, yolo\cite{yolo,yolov2,redmon2018yolov3,bochkovskiy2020yolov4} and other object detection works, as equation.\ref{eq:oldscale}:
\begin{equation}
\begin{split}
  s_k=s_{min}+\dfrac{s_{max}-s_{min}}{m-1}(k-1), k\in[1,m]
  \label{eq:oldscale}
\end{split}
\end{equation}
    where $s_{min}$ is $0.15$ and $s_{max}$ is $0.95$, meaning the lowest layer has a scale of $0.15$ and the highest layer has a scale of $0.95$, and $m$ is the index of feature maps.
    In this work, based on the analysis on characteristics of drone a progressive decay weight, $\beta_n$, is involved for scale in each layer. Thus, the scales in this work can be obtained as follow equation:
\begin{equation}
\begin{split}
  s_k=\beta_n(s_{min}+\dfrac{s_{max}-s_{min}}{m-1}(k-1)), k\in[1,m], n\in[1,m]
  \label{eq:newscale}
\end{split}
\end{equation}
    \par On the basis of statistics about drone sizes, features and position in different scales, we set anchor ratio including 1:2, 1:3, 2:1, 3:1 and 1:1 for different enhanced feature maps layers. Since it, we get a bunch of predict default boxes for different drone detection scenario.
    Additionally, we also utilize a series of data augmentation methodologies, extending the strategy in SSD\cite{liu2016ssd}. With probability of 3/5, the anchor-based sampling like data-anchorsampling in SSD\cite{liu2016ssd} is utilized, which randomly selects a drone in an image, crops sub-image containing the drone, and sets the size ratio between sub-image and selected drone to 512 multiply any within (0.1, 0.3, 0.5, 0.7, 0.9) randomly. For the remaining 2/5 probability, we adopt data augmentation with Gaussian Blur to simulate the drone in far or rapid motion and reduce the overfitting for specific kind of drone. Beside, we also inherit the data augmentation like horizontal flip and random color jitters in SSD\cite{liu2016ssd}, where the possibility of these two data augmentation are independent each other and with previous methods. We set Intersectionover-Union (IoU) threshold 0.5 to assign anchor to its ground-truth drones in order to enhance the recall rate of drones and guarantee anchor classification capability concurrently.
    In order to improve the recall rate of drones and ensure anchor classification ability simultaneously, we set IoU threshold 0.5 to assign anchor to its ground-truth drones.
\begin{table}[h]
    \centering
    \begin{tabular}{cccccc}
    \hline
        Feature & Stride & Size & Scale & Ratio & Parameter Number \\ \hline
        of\_1 & 4 & 128*128 & 28(56) & 1:(1,0.5,2) & 65536 \\ 
        of\_2 & 8 & 64*64 & 56(118) & 1:(1,0.5,1/3,2,3) & 24576 \\ 
        of\_3 & 16 & 32*32 & 118(176) & 1:(1,0.5,1/3,2,3) & 6144 \\ 
        of\_4 & 32 & 16*16 & 176(232) & 1:(1,0.5,1/3,2,3) & 1536 \\ 
        of\_5 & 64 & 8*8 & 232(326) & 1:(1,0.5,1/3,2,3) & 384 \\ 
        of\_6 & 128 & 4*4 & 326(408) & 1:(1,0.5,1/3,2,3) & 96 \\ 
        of\_7 & 256 & 2*2 & 408(484) & 1:(1,0.5,2) & 16 \\ 
        of\_8 & 512 & 1*1 & 484(526) & 1:(1,0.5,2) & 4 \\ \hline
    \end{tabular}
    \caption{The stride size, feature map size, anchor scale, ratio, and number of eight features maps for model.}
    \label{table:sacle}
\end{table}
    \subsection{Implementation Details}\label{Implementation Details}
    In this section, we start by outlining the specifics of how to construct our network. Although, the D-Linknet have published their save wight, the modifications operations on D-Linknet make the pertain model barely to use. However, as indications in ResNet\cite{restnet} it is still possible to achieve a performance better than pertained backbone model with a large dataset and training. Prior to model training processing, we implemented the pretrained ResNet on ImageNet to initializes a 'pretained model' with using the modified backbone networks and original architecture of SSD, in order to save iteration times. Then the 'xavier' method initializes all newly inserted convolution layer parameters. To improve our model, we utilize SGD with 0.9 momentum and 0.0005 weight decay. The batch size is 16 by default. For the beginning steps, the learning rate is set at $10^{-3}$, and decay to 0.8
    of current learning rate after 5 epochs.
    \subsection{Drone dataset construction}
    As pervious mentioned, each dataset has its advantages and disadvantages:
    Real World dataset has the most types of drones and backgrounds, but the resolution of the images is limited because they are collected from YouTube videos. Det-Fly dataset overcomes the limitations of single-view drone datasets by collecting drone images from other aircraft, which allows for comprehensive views of drones in different angles. However, it only includes one type of drone, which limits the model's ability to detect other types of drones. Similarly, MIDGARD and USC-Drone datasets also have limitations due to their single-type and single-view nature. Therefore, constructing a mixed dataset that combines images from multiple sources can provide a more diverse and comprehensive set of images that can improve the performance of drone detection models. Additionally, including other air objects such as birds in the dataset can also help researchers distinguish between drones and birds in long-distance detection scenarios.
    \par Overall, constructing a mixed dataset of drone images is crucial for improving the accuracy and reliability of drone detection models in real-world scenarios. In order to provide a more comprehensive and diverse range of images for training and testing drone detection models. A united dataset of drone images is constructed in this work to overcome the limitations of single-view and single-type drone datasets. The constructed dataset inherent the train and test set from Real World and USC-Drone dataset. For those datasets have not published their validation or test set, 10\% of published dataset will be selected as validation set and remaining counterparts as training set. In order to evaluate the robustness of drone detection model, we accept the sort label in MIDGARD dataset and concept in \cite{Det-Fly} to analyze the performance of model in different circumstances. In this work, the detect indoor drone is consider as  most difficult, followed by detecting drone outdoor within urban and the drone detection in countryside should be easiest. 
    \section{Evaluation}
     In order to get the comprehensive understanding of the unique challenges of drone detection due to factors such as varying views, scales, distances, and backgrounds. In this work, we invoke a number of datasets specifically designed for drone detection to train and verify the performance of models under realistic conditions.Real World dataset consists of images collected from YouTube videos and although it has limited resolution, it includes the most types of drones and backgrounds. Det-Fly dataset overcomes the limitation of single-view data by including multiple drone poses captured by cameras positioned at different angles. However, its usefulness is limited to detecting only one type of drone. Other datasets such as MIDGARD and Drone-vs-Bird also suffer from limitations of having a single type of drone and relatively limited view angles. Nonetheless, these datasets are essential tools for improving the accuracy and effectiveness of drone detection algorithms.
     Besides, in order to verify the contributions from each components and strategy in MDDPE, the constructed dataset is used for the thorough experimentation and ablation research. The comparison experiment between MDDPE and other state-of-art algorithm is also implemented to prove the superiority of MDDPE. Further more, robustness test for environment variation and types of drone is also involved to verify the performance of algorithms deeply.  
    \par
    As previous mentioned, the coco evaluation standard, including mAP, miss rate, false positive rate and precision-recall curves is invoked in this section to evaluate the performance of an object detection algorithm. To calculate mAP, the algorithm's predicted bounding boxes are compared to the ground truth bounding boxes using a measure called intersection over union (IOU). The formula for IoU is:
    $$IoU = \frac{Area(Prediction \cap Ground Truth)}{Area(Prediction \cup Ground Truth)}$$
    where $Area$ denotes the area of the bounding box.
    If the IOU between a predicted box and a ground truth box is above a certain threshold (0.5), then the prediction is considered a true positive. Thus, the formula for mAP can be expressed:
    $$mAP = \frac{1}{n}\sum_{i=1}^n AP_i$$
    where $n$ is the number of object classes, and $AP_i$ is the average precision of class $i$.

    \subsection{Backbones and input sizes experiments}
    To better understand the effect of backbone and input sizes, a serials of experiments are conducted to examine how different backbones and input sizes affect drone detection performance. Specifically, with using the same setting except for the feature extraction network in SSD\cite{liu2016ssd}, ResNet50, ResNet101 and modified D-linknet are implemented in this work. During the experiments, both of backbone model didn't use the pertain weight on ImageNet and train on real world dataset for 20 epochs. From Table \ref{table:backbone}, the proposed backbone with $512*512$ input size got 76.1\% on average, whereas the Resnet-50 gets 51.6\% and 61.05\% with 300*300 input size and 512*512 input size sequentially. Both Resnet-101 didn't show a better performance even with a deeper framework. The data indicate that more complexity model and higher Top-1 ImageNet classification accuracy may not benefit drone detection Map but the larger input size or feature maps can increase the performance directly. Therefore, in drone detection algorithm's framework, better performance classification are not necessary for better performance on detection. The modified D-linknet performs a reasonable speed benefited from the framework inherited from SSD and simple feature maps enhancement methods. For proposed framework, it runs 21 FPS on NVIDA GPU 2080ti during inference.
    \begin{table}[h]
    \centering
    \begin{tabular}{cccc}
    \hline
        Backbone & Input size & Parameters & $AP_{50}$ \\ \hline
        Resnet-50 & 300*300 & 8732 & 0.516 \\ 
        Resnet-101 & 300*300 & 8732 & 0.4178 \\  \hline
        Resnet-50 & 512*512 & 23290 & 0.6105 \\ 
        Resnet-101 & 512*512 & 23290 & 0.542 \\ 
        Modified D-Linknet & 512*512 & 98292 & 0.761 \\ \hline
    \end{tabular}
    \caption{The experiment results of the AP contributions from different backbone and input sizes.}
    \label{table:backbone}
\end{table}
    \subsection{Components Analysis on MDDPE}
    In this section, the detailed experiments and ablation studies are carried on the created dataset to assess the efficacy of a number of contributions made by proposed framework, such as the modified backbone, feature maps supplement function, feature maps recombination enhancement, and tailored anchors matching portfolio. With the exception of the noted component modifications, we maintain the same parameter values throughout all tests to allow for fair comparisons. On the created training set, all models are trained, and the validation set is used to assess them. Several baselines are defined and used to ablate each component's impact on the final performance in order to better comprehend the MDDPE. In the meantime, the SSD-512 is also invoked with using pertained Resnet-50 as a comparison to evaluate our work more comprehensively. 
    \subsubsection{Feature Maps Supplement Function}
     To do classification and regression at this step, we use an anchor designed in SSD \cite{liu2016ssd} and eight original feature maps produced using modified D-Linknet, named as M-D-Linknet as the baseline. Then, we install the last enhancement module and compare it to the basic M-D-Linknet-based SSD. Table \ref{table:component} shows that our feature maps supplement module, FMS, can improve M-D-Linknet-based SSD from 33.1\%, 76.1\%, 23.92\% to 36.56\%, 80.34\%, 26.52\% in $AP_{0.50:0.95}$, $AP_{0.50}$ and $AP_{0.75}$ respectively with using the feature map extracted from backbone. It is noteworthy that the feature maps supplement module has advanced improvements in the identification of tiny drones.

    \subsubsection{Feature maps recombination enhancement}
    Second, we use M-D-Linknet as the baseline to feature maps recombination enhancement function. We use the proposals methodology to enhance the eight feature maps from bottom to up and up to bottom. At this stage, we do not perform layer normalization before the feature map sent to predict layers. Table \ref{table:component} shows our feature maps recombination enhancement concept improve M-D-Linknet using FMRE from  36.56\%, 80.34\%, 26.52\% to 39.57\%, 84.25\%, 31.21\% in $AP_{0.50:0.95}$, $AP_{0.50}$ and $AP_{0.75}$ respectively. Analyzing in scale aspect, the FMRE has significant improvement on middle and larger scale drone performance.
    \subsubsection{Tailored anchors matching portfolio}
    To evaluate our tailored anchor matching portfolio, we use M-D-Linknet without anchor compensation as the baseline. Table\ref{table:component} shows that our improved anchor matching portfolio(TAMP) can improve M-D-Linknet with FMS, FMRE from 39.57\%, 84.25\%, 31.21\% to 47.76\%, 86.26\%, 44.53\%. in $AP_{0.50:0.95}$, $AP_{0.50}$ and $AP_{0.75}$ respectively. By analysis in recall statistic, the increase on recall statistics prove that our TAMP can give a comprehensive learning on all kinds of drone feature. 

    \begin{table}\resizebox{\linewidth}{!}{
    \centering
    \begin{tabular}{ccccccccccccc}
    \hline
        Componment & $AP_{0.50:0.95}$ & $AP_{0.50}$ & $AP_{0.75}$ & $AP_S$ & $AP_M$ & $AP_L$ & $AR_{0.50:0.95}$ & $AR_{0.50}$ & $AR_{0.75}$ & $AR_S$ & $AR_M$ & $AR_L$  \\ \hline
        SSD-512 & 0.3633  & 0.795 & 0.2844 & 0.1757 & 0.3824 & 0.4818 & 0.4338 & 0.4852 & 0.4988  & 0.3485 & 0.3514 & 0.5181  \\ 
        M-D-linknet & 0.331 & 0.761 & 0.2392 & 0.2087 & 0.3894 & 0.4312 & 0.3825 & 0.4476 & 0.4479 & 0.3485 & 0.4436 & 0.581  \\ 
        M-D-linknet + FMS & 0.3656 & 0.8034 & 0.2652 & \textbf{0.2235}  & 0.3648 & 0.5163 & 0.4046 & 0.4604 & 0.4613 & 0.3943 & 0.4159 & 0.6042  \\ 
        M-D-linknet + FMS +FMRE & 0.3957 & 0.8425 & 0.3121 & 0.2081 & 0.4175 & 0.5711 & 0.4212 & 0.4794 & 0.4808 & 0.3512 & 0.4865 & 0.6399  \\ 
        M-D-linknet + FMS +FMRE + TAMP & \textbf{0.4676} & \textbf{0.8626} & \textbf{0.4453} & 0.2126 & \textbf{0.5104} & \textbf{0.6213} & 0.5257 & 0.5581 & 0.5592 & 0.3199 & 0.5926 & 0.6871  \\ \hline
    \end{tabular}}
    \caption{The effectiveness of different components on the AP performance}
    \label{table:component}
\end{table}
    \subsection{Comparisons with State-of-the-Art}
    In order to estimate the robustness and generalization ability of our work, it is essential to test a drone detection model in different datasets due to the various factors that can affect the performance of the model. As previous mentioned, factors such as shot view, scale, distance, and background of detecting drones are diverse, thus making it necessary to have datasets with different types of drones and backgrounds. Real World dataset, for instance, comprises the most types of drones and environments, providing a more comprehensive view of the problem being studied. The dataset also includes limited-resolution images taken from YouTube videos, which presents an opportunity to study the limitations of drone detection under different viewpoints. Det-Fly dataset overcomes the shortcomings of the single-viewpoint drone data by incorporating multiple pose angles in its images. However, the dataset only contains one type of drone, limiting the model's applicability to other types of drones. Similarly, MIDGARD and Drone-vs-Bird datasets only contain one type of drone and a relatively rich environment but suffer from the same limitation of having a single viewpoint.
    \begin{table}[h]
    \centering
    \begin{tabular}{cccc}
    \hline
        ~ & real world &  det-fly & MIDGARD \\ \hline
        ours & \textbf{\textcolor{red}{0.801}} & \textbf{\textcolor{red}{0.977}} & \textbf{\textcolor{red}{0.9078}}  \\ 
        SSD512 & 0.761 & \textcolor{orange}{0.891} & 0.787  \\ 
         Yolov3 & - & \textcolor{blue}{0.877} & 0.723  \\ 
        Grid R-CNN & - & 0.824 & \textcolor{orange}{0.901}  \\ 
        FPN & - &  0.787 & 0.858  \\ 
        Faster R-CNN & - &  0.805 &  \textcolor{blue}{0.891}  \\ 
        RetinaNet & - & 0.779 & 0.888  \\ 
        Cascade R-CNN & - & 0.794 & 0.894 \\ \hline
    \end{tabular}
    \caption{The performance of different algorithm on four datasets with highest marked on red, second marked on orange and third marked on blue.}
    \label{table:differentdatset}
\end{table}
    \par Therefore, we test our drone detection model on different datasets to evaluate its performance under different scenarios, verify its accuracy and applicability. The table.\ref{table:differentdatset} show that the our framework has best performance in all drone datasets with 80.1\% in real world, 97.7\% in Det-fly and 90.78\% in MIDGARD. Moreover with inheriting framework of SSD\cite{liu2016ssd} our work still maintain a reasonable interface speech about 21 FPS on NVIDA GPU 2080ti. It is notably that due to the time limitations, we have recovered all mentioned methods in Table \ref{table:differentdatset} the data of some methods is referred from det-fly \cite{Det-Fly}. Although the USC-drone dataset is involved in this work, the officials labels published by author using one label file to label continuous motion of the same drone over 15 frames, where the position of some drones have been changed, which make this dataset currently barely use to evaluate.

\subsection{Robustness of detection scenario}
\begin{figure}[h]
  \centering  
  \includegraphics[width=\linewidth]{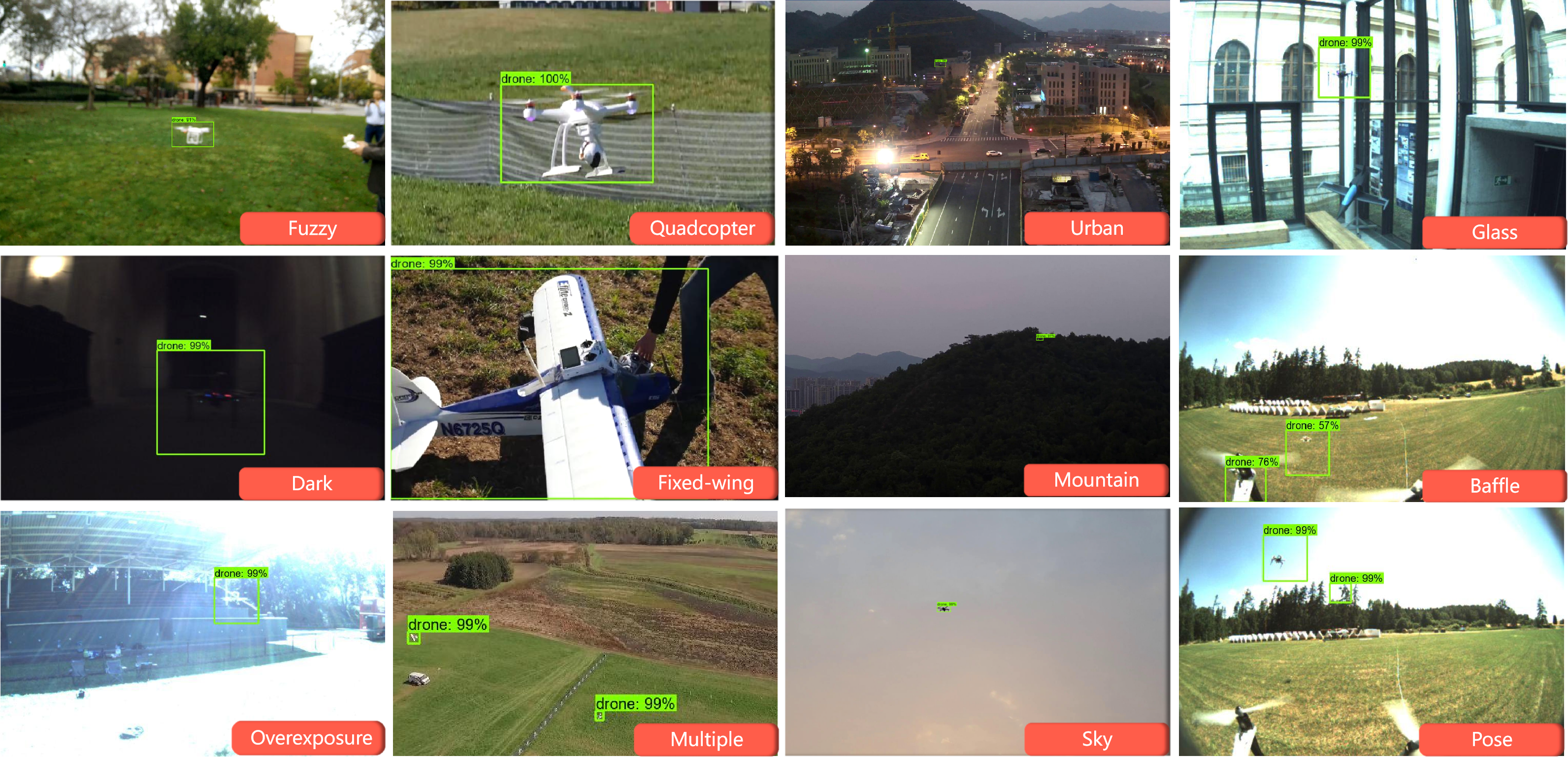} 
  \caption{The drone detection results on different kinds of detection scenario}  
  \label{fig:scenario}  
\end{figure}

    In order to evaluate the robustness of our model, the widely experiments on different drone detection scenario to simulate the realistic scenario is conducted. Since the sort label that proposed in \cite{Det-Fly} has not been published and the time limitations of this project, several examples of different drone detection scenario are selected manually to verify the robustness of MMDPE instead of the widely comparison between MMDPE and State-of-the-Art algorithms in this section. The Fig.\ref{fig:scenario} and \ref{moreexamples} show examples to demonstrate the effects of the MMDPE on handling drones under various scenario.
    \par
\begin{figure}[h]
    \centering  
    \includegraphics[angle=-90,width=\linewidth,]{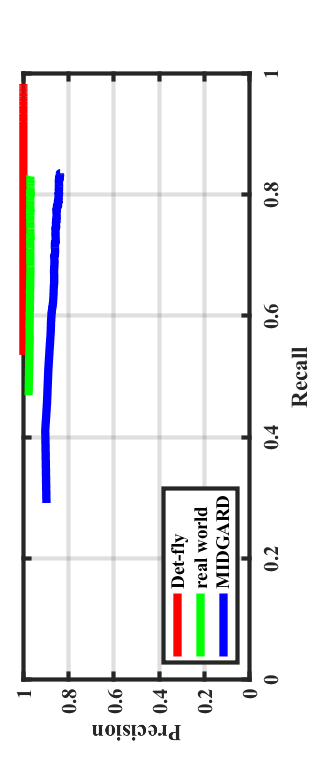} 
    \caption{The Precision-Recall curve of MDDPE in different datasets}
    \label{fig:pr} 
    \end{figure}

 As supplementary, the Fig.\ref{fig:pr} demonstrate the Precision-Recall curve of our MMDPE in different datasets. Although the more complex drone detection environment in MIDGARD slightly declines the detection confidence of MDDPE, it is obvious that the MDDPE is able to handle the various challenges in shot view, scale, distance, and background  from datasets.
    
     All experiment proves that our drone detection model have a strong Robustness on both environment, scenario and type of drones.
    \section{Conclusion}
    This paper introduces a novel drone detection algorithm with multiple pyramid feature maps enhancement structure (MDDPE). In this work, novel feature maps enhance module including, feature maps supplement function and feature maps recombination enhancement function that utilizes different level information are proposed and thus obtains more discriminability and robustness features. Auxiliary supervisions introduced in early layers by using tailored anchors are adopted to effectively facilitate the drone features. Moreover, an improved anchor matching strategy is introduced to match anchors and ground truth drone as far as possible to provide better real drone detection scenario  simulation and initialization for the regressor. Comprehensive experiments are conducted on popular drone detection benchmarks, to demonstrate the superiority of proposed MDDPE compared with the state-of-the-art detectors. For future research, the attention and modified feature maps recombination enhancement function is planned to implement to improve the performance of model, especially for small scale drone detection. In addition, more state-of-the-art detectors and sorting methodology for different detection scenario will be implemented for comprehensively comparison.
    \newpage
    \appendix
    \bibliographystyle{IEEEtran}
    \bibliography{paper}
    \addcontentsline{toc}{section}{Reference}
    \newpage
    \section{Appendix}
    \subsection{more examples for MDDPE}\label{moreexamples}
    \begin{figure}[h!]
    \centering  
    \includegraphics[width=\linewidth,]{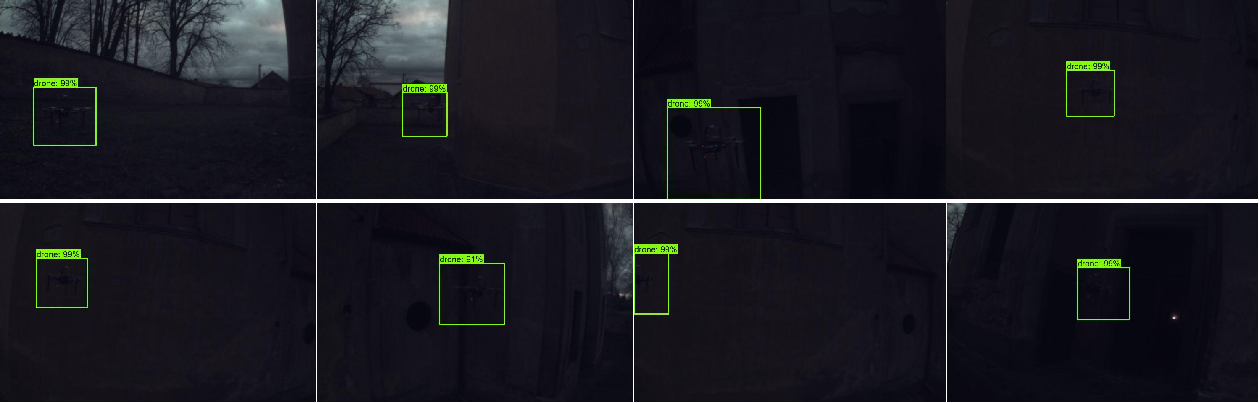} 
    \caption{The results of MDDPE in dark scenario}
    \end{figure} 
    \begin{figure}[h!]
    \centering  
    \includegraphics[width=\linewidth,]{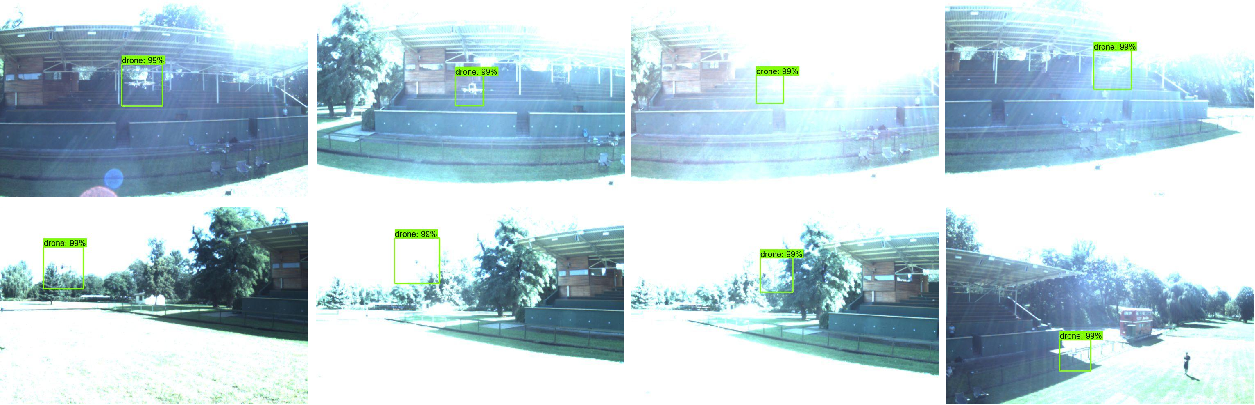} 
    \caption{The results of MDDPE in over exposure scenario}
    \end{figure} 
    \begin{figure}[h!]
    \centering  
    \includegraphics[width=\linewidth,]{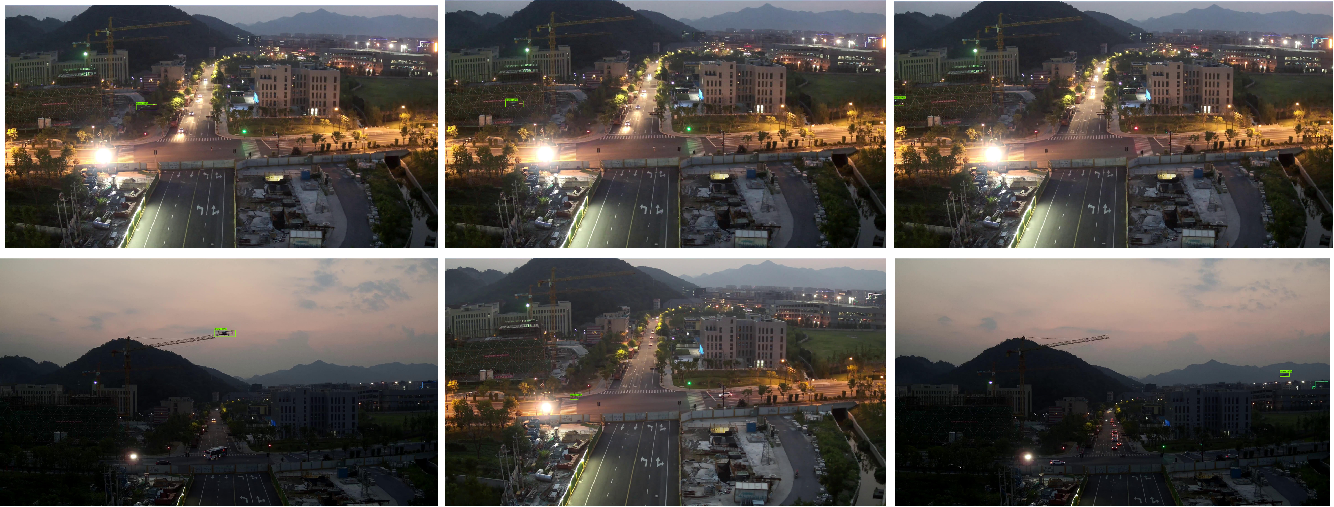} 
    \caption{The results of MDDPE in urban scenario}
    \end{figure} 
\end{document}